# Artificial Intelligence Assisted Infrastructure Assessment Using Mixed Reality Systems

**Enes Karaaslan**
Ph.D. Candidate, Department of Civil, Environmental, and Construction Engineering, University of Central Florida, 12800 Pegasus Drive, Orlando, Florida 32816, USA, TEL: 321-975-9775, Email: Enes.Karaaslan@ucf.edu

**Ulas Bagci**
Professor, Center for Research in Computer Vision (CRCV), University of Central Florida, 4328 Scorpius Street, Orlando, Florida 32816, USA, TEL: 407-823-1047, Fax: 407-823-0594, Email: bagci@ucf.edu

**F. Necati Catbas**
Professor, Department of Civil, Environmental, and Construction Engineering, University of Central Florida, 12800 Pegasus Drive, Orlando, Florida 32816, USA, TEL: 407-823-2841, Fax: 407-823-3315, Email: catbas@ucf.edu (corresponding author)

Word count:  5,240 word texts + 3 tables x 250 words (each) = 5,990 words

Submission Date: 08/01/2018

# 1. ABSTRACT

Conventional methods for visual assessment of civil infrastructures have certain limitations, such as subjectivity of the collected data, long inspection time, and high cost of labor. Although some new technologies (i.e. robotic techniques) that are currently in practice can collect objective, quantified data, the inspector's own expertise is still critical in many instances since these technologies are not designed to work interactively with human inspector. This study aims to create a smart, human-centered method that offers significant contributions to infrastructure inspection, maintenance, management practice, and safety for the bridge owners. By developing a smart Mixed Reality (MR) framework, which can be integrated into a wearable holographic headset device, a bridge inspector, for example, can automatically analyze a certain defect such as a crack that he or she sees on an element, display its dimension information in real-time along with the condition state. Such systems can potentially decrease the time and cost of infrastructure inspections by accelerating essential tasks of the inspector such as defect measurement, condition assessment and data processing to management systems. The human centered artificial intelligence (AI) will help the inspector collect more quantified and objective data while incorporating inspector's professional judgement. This study explains in detail the described system and related methodologies of implementing attention guided semi-supervised deep learning into mixed reality technology, which interacts with the human inspector during assessment. Thereby, the inspector and the AI will collaborate/communicate for improved visual inspection.

# 2. INTRODUCTION

Federal Highway Administration provide annual statistics on structurally deficient bridges. According to 2017 statistics, 54,560 bridges are structurally deficient among total number of 54,000 bridges (*1*). Utilizing novel technologies for better management of such aged and deteriorated civil infrastructures is becoming more critical. While the existing status of the US civil infrastructure is well documented, there is slow progress in improving this status. Structural systems have aged to an extent that critical decisions such as repair or replacement should to be made effectively. To prevent the impending degradation of civil infrastructure, utilizing novel technologies for periodic inspection and assessment for long term monitoring has recently become more critical (*2*). Although the inclination to use conventional inspection methods still persists, advanced sensing technologies have the ability to better understand the current condition with more resolution and accuracy (*3*). Conventional methods for visual assessment of infrastructures have certain limitations, such as subjectivity of the collected data, long inspection time, and high cost of labor. On the other hand, imaging technologies allow collecting quantified data and performing objective condition assessment. These techniques are now receiving a breakthrough improvement with the employment of the state-of-the-art Artificial Intelligence (AI) models. Instead of post-processing of the collected inspection data, an AI system can detect the damages in real-time and analyze for condition assessment at a reasonable accuracy. The main objective of the AI integrated Mixed Reality (MR) system described in this paper is to assist the inspector by accelerating certain routine tasks such as measuring all cracks in a defect region or calculating a spall area. In this system, the human-centered AI interacts with the inspector instead of completely replacing the human involvement during the inspection. This collective work will lead to quantified assessment, reduced labor time while also ensuring human verified results. Even though this study focused on concrete defect assessments with particular focus on concrete bridges, the methodology can be expanded for other types of structures.

## 2.1. Virtual, Augmented and Mixed Reality

Virtual Reality (VR) is a computer simulated reality that replicates a physical environment or imaginary world through an immersive technology. VR replaces the user's physical world with a completely virtual environment and isolates the user's sensory receptors (eyes and ears) from the real world (*4*). The VR is observed through a system that displays the objects and allows interaction, thus creating virtual presence (*5*). Nowadays, VR headsets have gained vast popularity especially in gaming industry. The Augmented Reality (AR), on the other hand, is an integrated technique that often leverages image processing, real-time computing, motion tracking, pattern recognition, image projection and feature extraction. It overlays computer generated content onto the real world. An AR system combines real and virtual objects in a real environment by registering virtual objects to the real objects interactively in real time (*6*). The beginning of AR dates back to Ivan Sutherland's see-through head-mounted display to view 3D virtual objects (*7*). The initial prototype was only able to render few small line objects. Yet, the AR research has recently gained dramatic increase and now it is possible to visualize very complex virtual objects in the augmented environment. The recent developments of AR/VR technology helped companies produce holographic headsets that benefits Mixed Reality (MR) technology, in which one can experience hybrid reality where physical and digital objects co-exist and interact in real time. The term Mixed Reality was originally introduced in a 1994 paper "A Taxonomy of Mixed Reality Visual Displays" (*8*). In the paper, a Virtuality Continuum (VC), in other words, mixed reality spectrum was explained in detail. A schematic representation is shown in Figure 1.

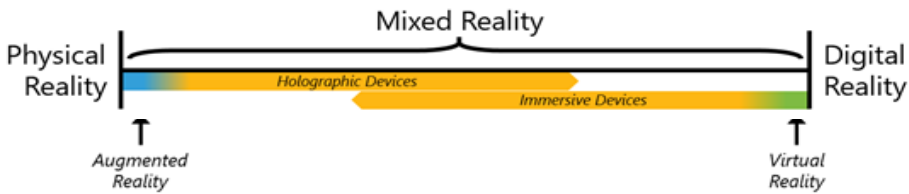

**Figure 1 Mixed reality spectrum and device technologies** (*9*)

The MR technology has breakthrough applications especially with successful deployment of 3D user interfaces such as in computer-aided design, radiation therapy, surgical simulation and data visualization (*10*). The next generation of computer games, mobile devices, and desktop applications also will feature 3D interaction (*11*). There are also some other efforts for using MR technology in construction industry and maintenance operations. Kamat and El-Tawil (2007) discusses the feasibility of using AR to evaluate earthquake-induced building damage. Behzadan and Kamat (2007) investigated the application of the global positioning system and 3 degree-of-freedom (3-DOF) angular tracking to address the registration problem during interactive visualization of construction graphics in outdoor AR environments (*13*). The vision-based mobile AR systems are vastly used in 3D reconstruction of scenes for architectural, engineering, construction and facility management applications. Bae et al. (2013) developed a context-aware AR system that generates 3D reconstruction from 3D point cloud. Important effort for use of AR in infrastructure inspection is also shown by several researchers (*14*). Researchers in University of Cambridge currently collaborate with Microsoft to develop an effective bridge inspection practice in which the data collected from the field is visualized in MR environment in the office (*15*). Moreu et al. (2017) developed a conceptual design for novel structural inspection tools for structural inspection applications based on HoloLens (*17*) device (*16*). The experiments conducted with the HoloLens for taking measurements and benchmarking the obtained measurements are shown in the study. The proposed methodology takes even a further step and combines AI implementation with MR technology. In this system, the embedded AI architecture not only predicts the location/region of cracks and spalling on the infrastructure in real-time along with condition information but also augments the information in the holographic headset for improved human inspector - AI interaction.

## 2.2. Overview of Deep Learning Approaches in Damage Detection and Analysis

For more than a decade, researchers have been investigating employing the techniques in the Computer Vision field to analyze cracks, spalls and other types of damages. The early approaches mostly used edge detection, segmentation and morphology operations. Yet, the recent advances in AI yielded very promising accuracy and possessed a wide range of applicability. A review paper on computer vision based defect detection and condition assessment of concrete infrastructures emphasizes the importance of sufficiently large, publically available and standardized datasets to leverage the power of existing supervised machine learning methods for damage detection (*18*). According to the study, learning based methods can be reliably used for defect assessment. For the processing of defect images, many researchers in the literature implemented Convolutional Neural Network (CNN) to perform automatic crack detection on concrete surfaces. Combined with transfer learning and data augmentation, CNN can offer highly accurate input for structural assessment. Yokoyama and Matsumoto (2017) developed a CNN based crack detector with 2000 training images (*19*). The main challenge of the detector was that the system often classifies stains as cracks. Yet, the detection is successful for even very minor cracks. Similarly, Jahanshahi and Masri (2012) developed a crack detection algorithm that however uses an adaptive method from 3D reconstructed scenes (*20*). The algorithm extracts the whole crack from its background, where the regular edge detection based approaches just segment the crack edges; thereby offering a more feasible solution for crack thickness identification. Adhikari et al. (2014) used 3D visualization of crack density by projecting digital images and neural network models to predict crack depth, necessary information for condition assessment of concrete components (*21*).

For detection of spalls and cracks, German et al. (2012) used entropy-based thresholding algorithm in conjunction with image processing methods in template matching and morphological operations (*22*). In addition to detection of local defects of structures, there are also studies on identifying global damages of the structures. Zaurin et al. (2015) performed motion tracking algorithms to measure the mid-span deflections of bridges under the live traffic load (*23*). Computer Vision is also used to process ground penetration radar (GPR) and infrared thermography (IRT) images that are useful to identify delamination formed inside the concrete structures. Hiasa et al. (2016) processed the IRT images of bridge decks taken with high-speed vehicles (*25*). In identifying damages, many different techniques are useful for specific purposes. However, a more generalized deep learning approach is introduced in this study so that the methods can be expanded toward identifying almost any type of damage if sufficient amount of training data is available.

The CNN models are mostly composed of convolutional and pooling layers. In the convolutional layers, the input images are multiplied by small distinct feature matrices that are attained from the input images (corners, edges etc.) and their summations are normalized by matrix size (i.e. kernel size). By convolving images, basically similarity scores between every region of the image and the distinct features are assigned. After convolution, the negative values of similarity in the image matrix are removed in the activation layer by using the rectified linear unit (ReLU) transformation operation. After the activation layer, the resultant image matrix is reduced to a very small size and added together to form a single vector in the pooling layer. This vector is then inserted in fully connected neural network where actual classification happens. The image vectors of the trained images are compared with the input image and a correspondence score is calculated for each classification label. The highest number will indicate the classified label. A summary of the described procedure is shown in Figure 2.

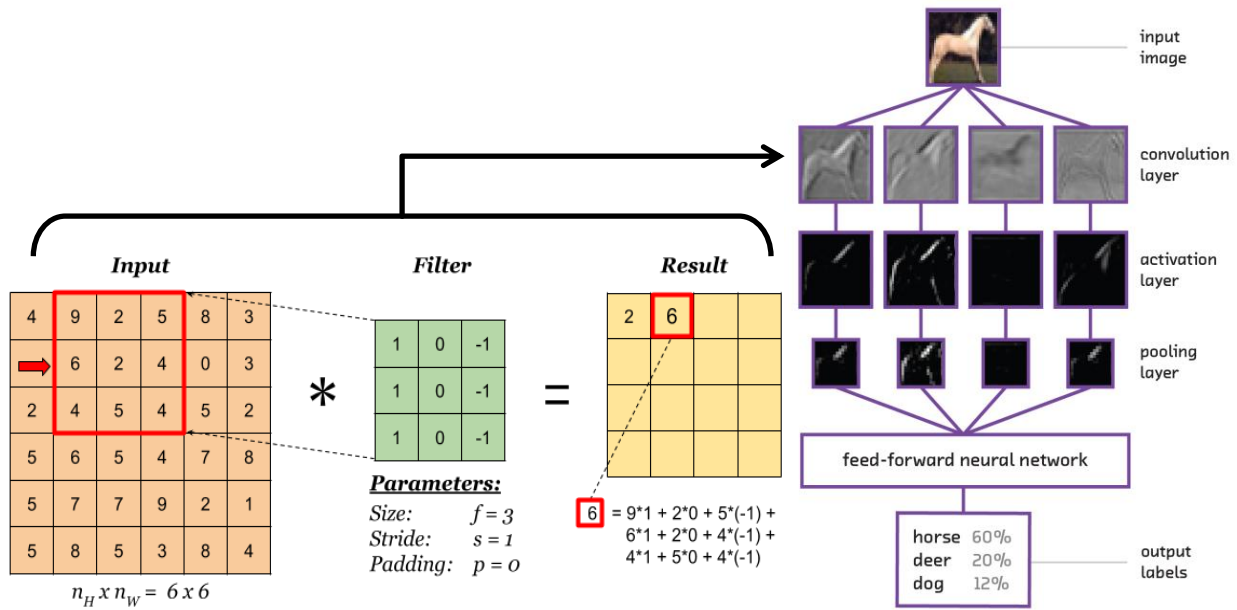

**Figure 2 Description of a simple convolutional neural network (CNN)**

## 3. METHODOLOGY

The proposed AI assisted infrastructure assessment using MR technology employs the state-of-the-art methods and algorithms from interdisciplinary practices. Machine learning is vastly used for robust detection of cracks and spalls on infrastructures whereas human-computer interaction concepts are employed for improving the assessment performance by including the professional judgement of the human inspector. MR is an excellent platform to maintain this interaction since it augments virtual information into the real environment and allows the user to alter the information in real-time. In this proposed methodology, bridge inspector uses MR headset during routine inspection of infrastructure. While the inspector performs routine inspection tasks, the AI system integrated into the headset continuously guides the inspector and shows possible defect locations. If a defect location is confirmed by the human inspector, the AI system starts analyzing it by first executing defect segmentation, then characterization to determine the specific type of the defect. If the defect boundaries need any correction or segmentation needs to be fine-tuned, the human inspector can intervene and calibrate the analysis. The alterations made by the human inspector (e.g. change of defect boundary, minimum predicted defect probability etc.) will be used later for retraining of the AI model by following a semi-supervised learning approach. Thereby, the accuracy of AI is improved over time as the inspector corrects the system.

Another advantage of the system is that the inspector can analyze defects in a remote location while reducing need for access equipment. Even though in some cases, hands-on access is evitable (i.e. determining subconcrete defects); the system can be still effective for quick assessments in the remote location. If the defect location is far or in a hard to reach location, the headset can zoom in and still perform assessment without needing any access equipment such as snooper truck or ladder. The proposed framework is illustrated in Figure 3.

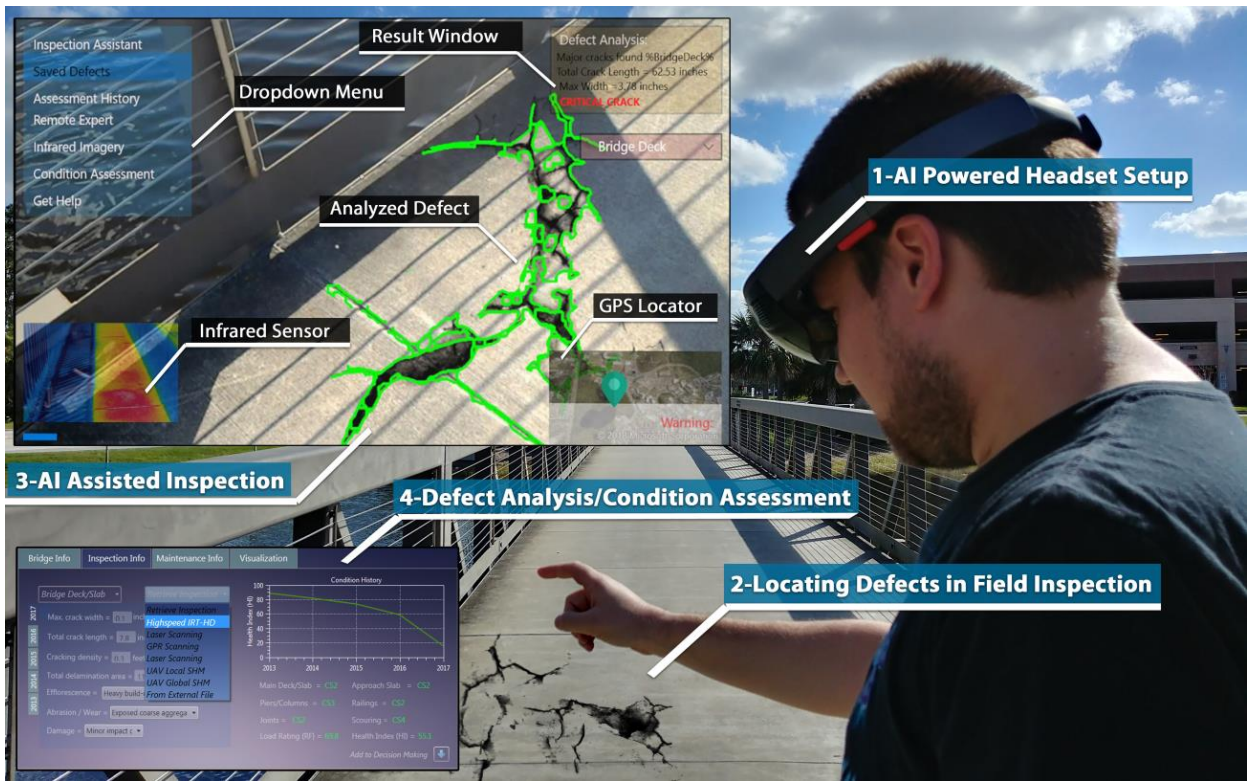

**Figure 3 Visual representation of the AI powered mixed reality system. (The headset user interface and analysis environment are shown for illustration purposes.)**

The proposed methodology of AI assisted infrastructure assessment using MR systems differs from the state-of–practice of current learning based approaches and mixed reality implementations in several aspects. Table 1 shows comparison of the proposed method with major literature work. The major difference of the proposed method from the current mixed reality approaches is that the system performs automatic detection and segmentation of the defect regions using real-time deep learning operations instead of manually marking the defect regions in the MR platform. In this way, the system can save significant amount of time in defect assessment as opposed to marking all these defects in the current MR implementations.

**Table 1 Comparison of the proposed research with the major literature work**

| Ioannis (2017) | Moreu et al. (2017) | Bae et al. (2013) | Xie et al. (2017) | Proposed Method |
|---|---|---|---|---|
| Remote bridge inspections with HoloLens | Structural inspection and measurement using HoloLens | Mixed reality for structure 3D scene reconstruction | CNN based crack detection | Mixed reality assisted bridge condition assessment |
| Data collections is monitored from an remote location | On-site measurement of defects | Image data is reconstructed after the data collection | Aims post processing of images to identify defects | On site system to augment bridge inspector's performance |
| Focuses on visualization and post-processing of data | Relies on human mostly while obtaining measurements | No detection of defects is implemented, 3D model is used for inspection | Detection performance relies on AI system only | Aims creating a collective intelligence with human - AI collaboration |
| Views high - resolution defect images on real size bridge model | Uses 3D projective geometry for measurement estimation | Uses widely 3D projective geometry to register images | Uses basic data augmentation techniques to increase training dataset | Uses an extensive data augmentation that generates many variations of defect images |

### 3.1. Data Collection Procedure and Defect Characteristic

Automated detection of defects in concrete structures requires training of each defect type individually, by processing a large number of training images. First, commonly available infrastructure defects are determined and their condition assessment procedure is investigated using the infrastructure inspection guides (*27–29*). According to the reference guides, the infrastructure defect types were determined as shown in Figure 4: a. Cracking, b. Rusting, c. Spalling, d. Efflorescence, e. Joint Damage, f. Delamination (detected by infrared).

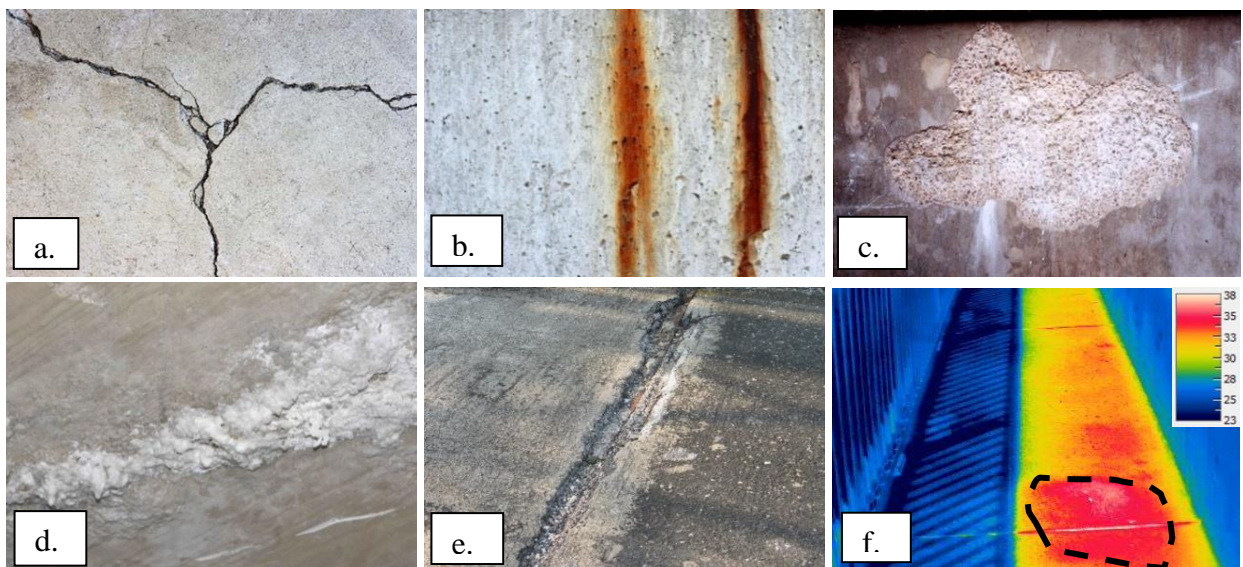

**Figure 4 Example defect images for each defect classification.**

Data collection is an important step of developing an AI system. The significant challenges of potential real life applications need to be evaluated carefully to collect suitable data for AI training. A preliminary work has been conducted in CITRS Lab (Civil Infrastructure Technologies for Resilience & Safety) at the University of Central Florida (UCF) in order to determine the important aspects of field data collection procedure. The effects of illumination, maximum crack width, target distance and camera resolution have been investigated in a laboratory environment. A set of synthetically generated crack images with different thicknesses, brightness and pattern are printed on letter size papers and placed on white platform. The experiment setup is demonstrated in Figure 5.

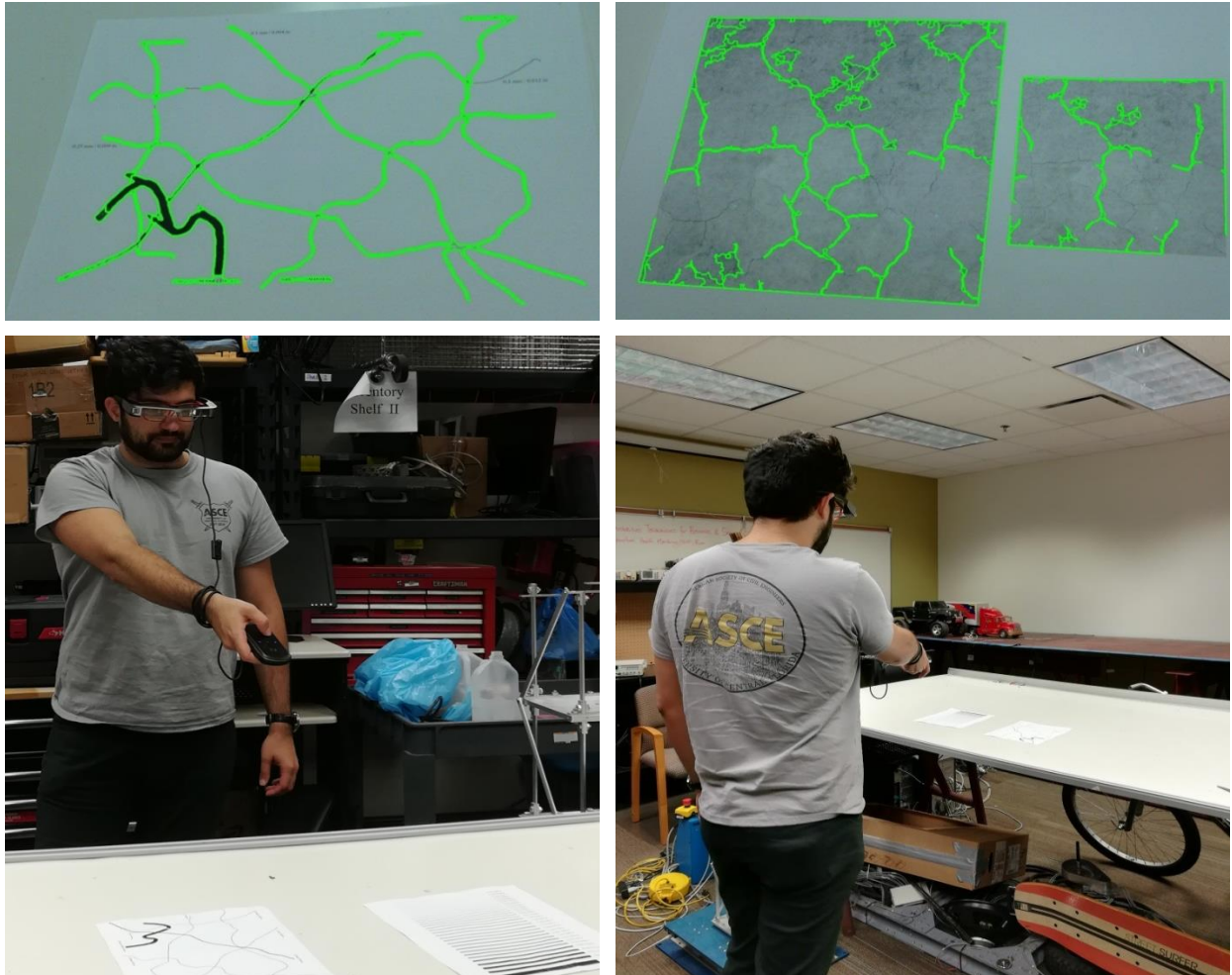

**Figure 5 Preliminary work on laboratory data collection held in EDM Lab.**

The proposed methodology will focus on cracks and spalls for this study and expand the system scope in the future with more defect types. The available defect images are gathered from various sources including industry partners, transportation agencies and other academic institutions. Some of the data were only categorized but not annotated; considerable portion of the data were annotated with bounding box pixel coordinates and a relatively small dataset was annotated for segmentation. An extensive data augmentation was however applied to the datasets to further increase AI prediction accuracy. The data augmentation included rotation, scaling, translation and Gaussian noise. The annotation styles of all of the training datasets were unified and converted to Pascal VOC 2012 annotation format. The summary information of the training datasets is shown in Table 2.

**Table 2 Summary of the training datasets**

| Dataset Annotation | Class Types | Dataset Size | Source |
|---|---|---|---|
| Sub-cropped, labeled but not annotated | Cracking and intact concrete | 40,000 images (with large data augmentation) | Concrete crack dataset (*30*) |
| Labeled and annotated for boundary boxes | Line crack, alligator crack, joint failure, spalling | 9000 images, 15500 labels (no data augmentation) | Road damage dataset (*31*) |
| Labeled and annotated for segmentation | Cracking and spalling | 2000 images (with little data augmentation) | Bridge inspection dataset (*32*) |
| Labeled and annotated for segmentation | Cracking and spalling | 300 images (with no data augmentation) | Image scrapping and some field data |

The trainings of the AI models were performed in the Newton Visualization Cluster operated by UCF Advanced Research Computing Center (2014). The Newton Visualization Cluster includes 10 compute nodes with 32 cores and 192GB memory in each node; two Nvidia V100 GPUs are available in each compute node totaling 320 cores and 20 GPUs. The model trainings were performed on two clusters with total of 4 GPUs. A single training was executed for 1million steps (Approximately takes 75 hours). The training was repeated for multiple times in order to find optimal hyperparameters.

### 3.2. Real Time Damage Detection

For real time detection of damages, a light weight architecture that can run on mobile CPUs was selected. SSD: Single Shot MultiBox Detector (SSD) is a relatively new, fast pipeline developed by Liu et al. (2016). SSD uses multi boxes in multiple layers of convolutional network and therefore has an accurate region proposal without requiring many extra feature layers. SSD predicts very fast while sacrificing very little accuracy, as opposed to other models in which significantly increased speed comes only at the cost of significantly decreased detection accuracy (*35*). The network architecture of the original SSD model is shown in Figure 6.

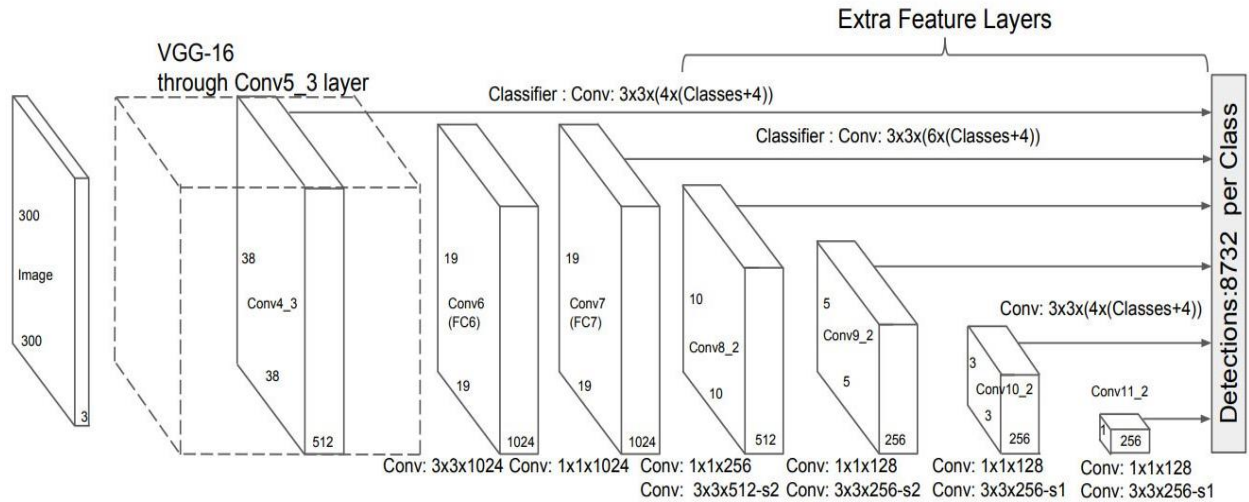

**Figure 6 Original SSD network architecture** (*34*).

Original SSD paper uses VGG-16 as a base architecture. VGG has become widely adopted classifier after it won the 2015 ImageNet competition (*36*). Although newer classifiers such as MobileNetV2 offers much faster prediction speeds at similar accuracy in a 15 times smaller network (*35*), VGG is a better choice to benefit transfer learning in this study (due to extensive hardware and memory requirement of MobileNETv2). Transfer learning allows employing the weights of already trained networks by fine-tuning only the certain classifier layers based on the size of the available dataset. Figure 7 shows challenging cases where damage detection algorithm from real-world images show promising results.

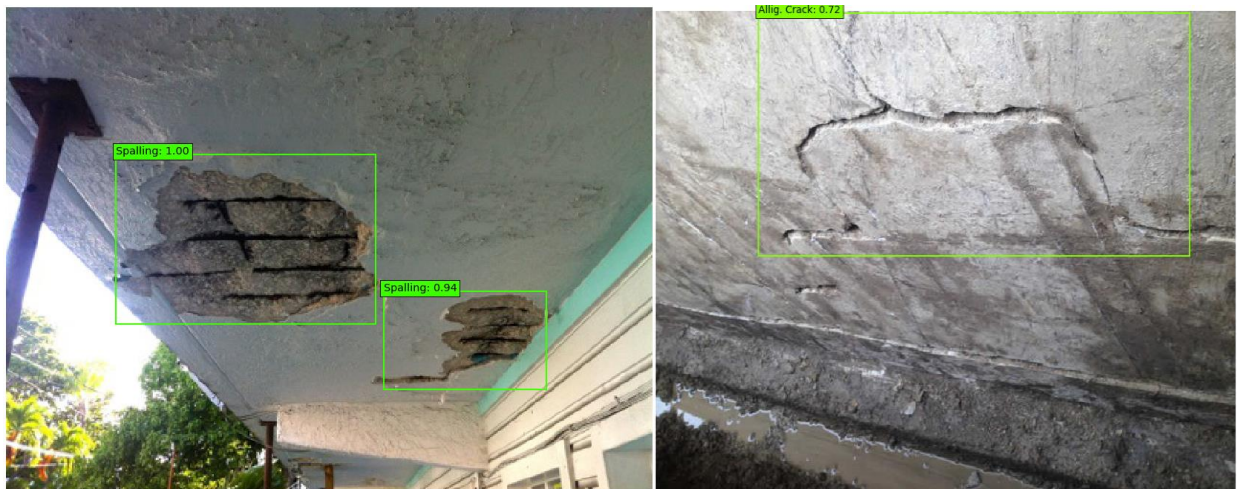

**Figure 7 Damage detections on real-world images (Left: Spalling in multi locations at different depth, Right: Alligator crack detected at large angle on wetted concrete surface).**

### 3.3. Attention Guided Segmentation

For concrete defect assessment, it is not solely enough to detect the damage in a bounding box; but the damage also needs to be segmented from intact regions in order to perform defect measurement. Therefore, another AI model is implemented in parallel to the SSD to perform segmentation of the damage regions. Popular segmentation models such as FCN, UNet, SegNet and SegCaps (*37*) are investigated; however their architectures are found to be too large for the small annotated dataset used in this study. To overcome this problem, the VGG weights that were re-trained in SSD architecture was used in a relatively small, customized segmentation architecture that is inspired by SegNet model (*38*). The SegNet model architecture is shown in Figure 8.

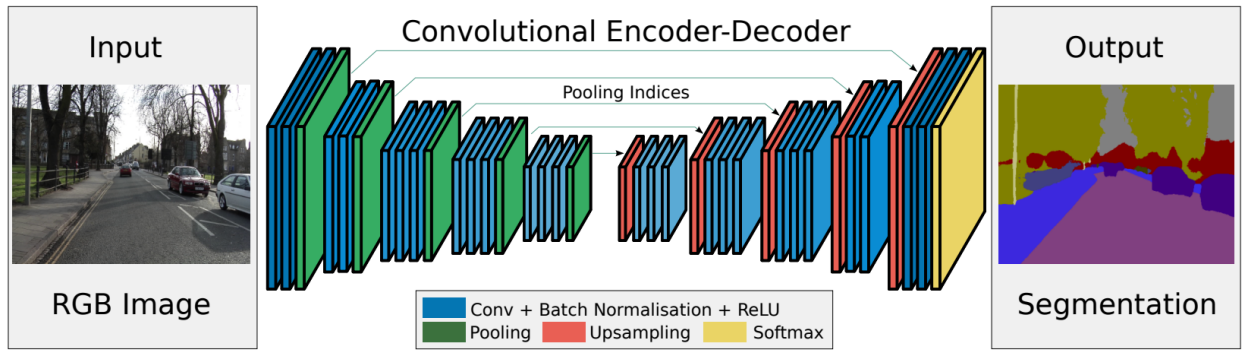

**Figure 8 An illustration of the *SegNet* architecture** (*38*)**.**

As a unique approach for damage segmentation, an attention guided technique is proposed in this paper. A sequential connection is created between detection and segmentation models. First, images are first fed into damage detection pipeline and when the bounding box is verified by the human-inspector, damage segmentation is executed only for the region inside the detected bounding box. This approach significantly improves the accuracy of segmentation and successfully prevents outliers. Figure 9 shows qualitatively how attention guided segmentation is superior to the segmentation without attention guidance. In the figure, the segmentation model is first executed for the entire image yielded inaccurately segmented regions. In the second image, only the bounding box region is fed into the segmentation pipeline and resulted in much higher accuracy.

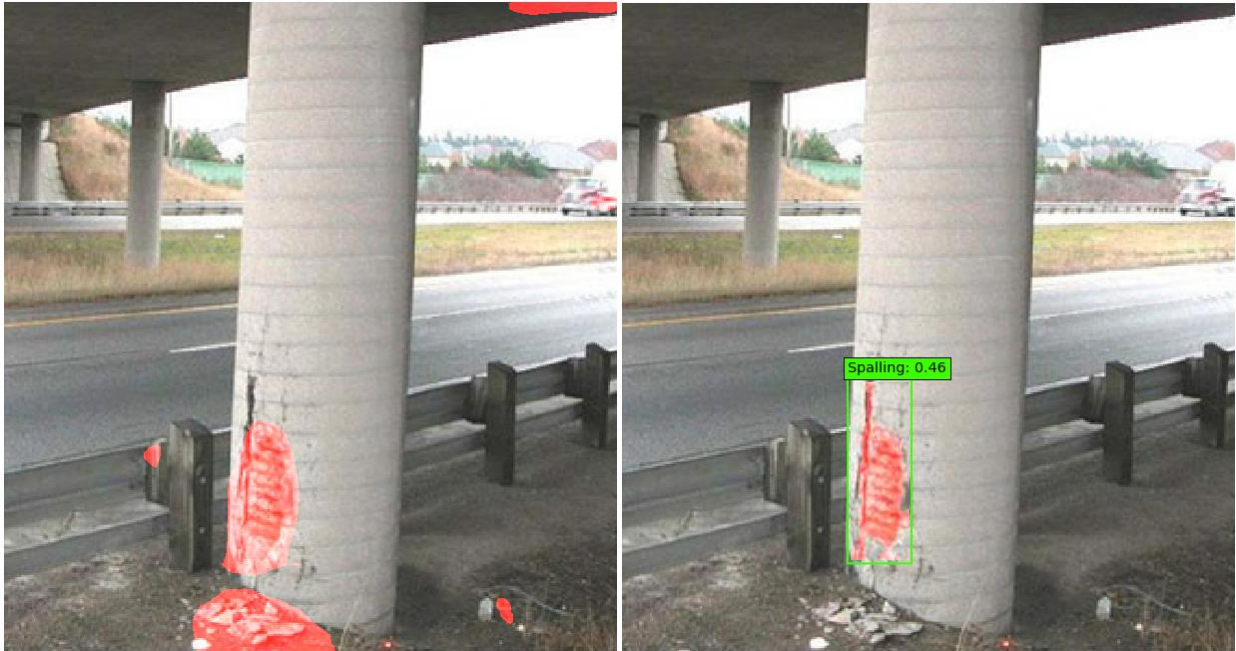

**Figure 9 Effectiveness of attention guided segmentation shown in red highlighted areas (Left: Segmentation resulted in some false positive results; Right: Attention guidance readily removes misclassified pixels).**

### 3.4. Human-Centered AI and Semi-supervised Learning

The human-computer interaction in mixed reality (MR) technology will allow benefitting human-AI collaboration for collective intelligence. In the proposed AI models for damage detection and segmentation, the prediction threshold values in the inference mode are adjusted by the human inspector thorough the MR system. This type of hybrid AI can easily outperform a traditional AI alone (*39*). This type of hybrid system is commonly seen in autonomous vehicle technologies, health industry and video game AI engines. When coupled with semi-supervised learning hybrid AI can perform impressively well.

During a bridge inspection, by asking the human inspector to modify prediction threshold will help improving the accuracy of the detection and determining the boundary region of the segmentation. In Figure 10, real-time damage detection is not showing one of the spall regions to the inspector when the prediction threshold is set to 0.5; when the inspector adjusts the value to 0.2, the missing spall region is also detected. (the value represents probability of accurate prediction.)

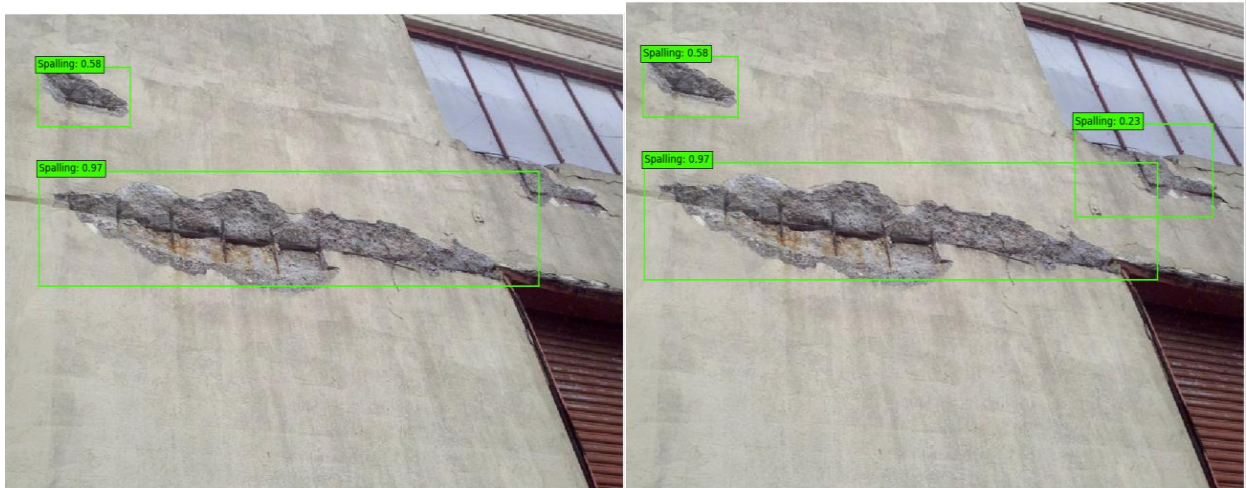

**Figure 10 Example of human-AI collaboration in the proposed methodology (detection AI alone on the left misses a spall, while human-assisted AI detects all spalls on the right with threshold adjustment by the inspector).**

Similarly, the human inspector can also fine-tune the segmentation boundary by adjusting the prediction threshold. Thus, the damage area can be calculated at higher accuracy. The fine-tuned segmentations along with the corresponding bounding box coordinates are recorded for future re-training while benefitting from semi-supervised learning. Some example results of human-AI collaborative damage detection and segmentation are shown in Figure 11.

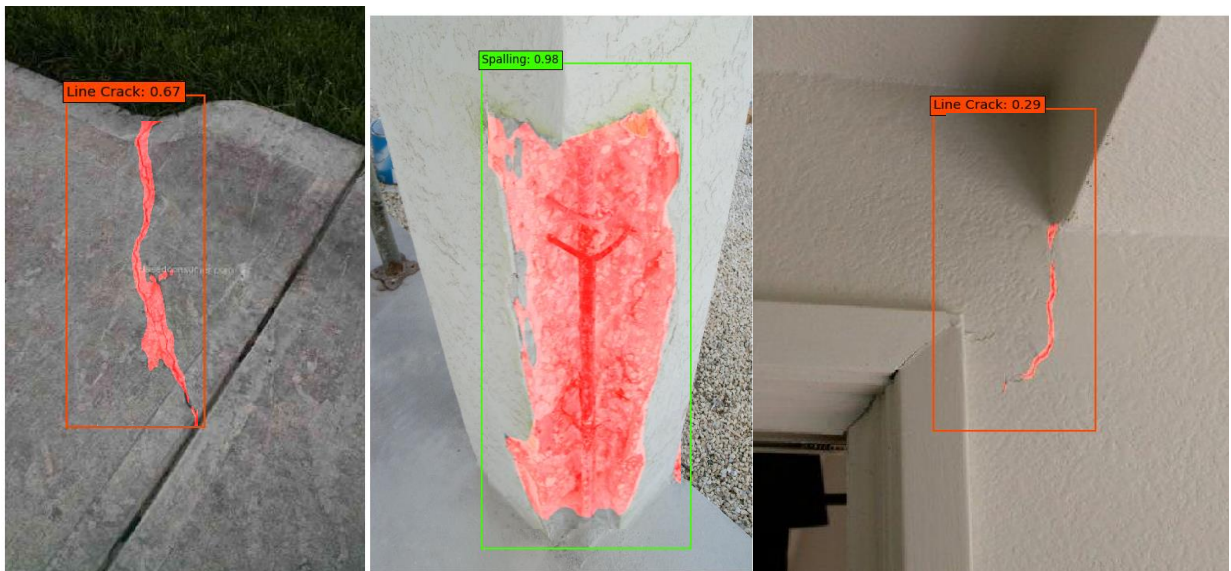

**Figure 11 Example results of human-AI collaborative damage detection and segmentation.**

### 3.5. Pose Estimation and Geometry Calculation

The condition assessment methodology based on the AI system's damage analysis will require answers to following: "how wide is this crack?", "Which one of the bridge piers is closer?", "What is the camera height, rotation or focal length?" This information is required for identifying actual measures of defects for accurate assessment of infrastructures and also for augmenting a certain object onto 3D view or highlighting defects in an MR headset. Using projective geometry and camera calibration models, it is possible to perform correct projections of objects onto 3D, achieve scene reconstruction and accurately predict actual dimension of the objects. However performing transformations in 3D spaces requires use of 4D projective geometry instead of conventional 3D Euclidian geometry (*40*). The projection matrix allowing camera rotation is defined as in Equation 1.

$$x = K[R \quad t]X \tag{1}$$

where; $x$: Image coordinates, $K$: Intrinsic matrix, $R$: Rotation matrix, $t$: Translation, $X$: World coordinates. The projected coordinate vector x is calculated by multiplying the world coordinates by the rotation and translation free projection matrix. The coordinate parameters are then put into system of equation as in (2.

$$w \begin{bmatrix} u \\ v \\ 1 \end{bmatrix} = K \begin{bmatrix} \alpha & s & u_0 \\ 0 & \beta & v_0 \\ 0 & 0 & 1 \end{bmatrix} \begin{bmatrix} r_{11} & r_{12} & r_{13} & t_x \\ r_{21} & r_{22} & r_{23} & t_y \\ r_{31} & r_{32} & r_{33} & t_z \end{bmatrix} \begin{bmatrix} x \\ y \\ z \\ 1 \end{bmatrix} \tag{2}$$

The local coordinates on image plane are represented by u and v; w defines the scale of the projected object. $\alpha$ and $\beta$ stand for rotation angles with respect to coordinate axes and s is short for sinus function. Unity allows camera control that help developers perform correct projection onto image plane form 3D view. The projection is described in Figure 12.

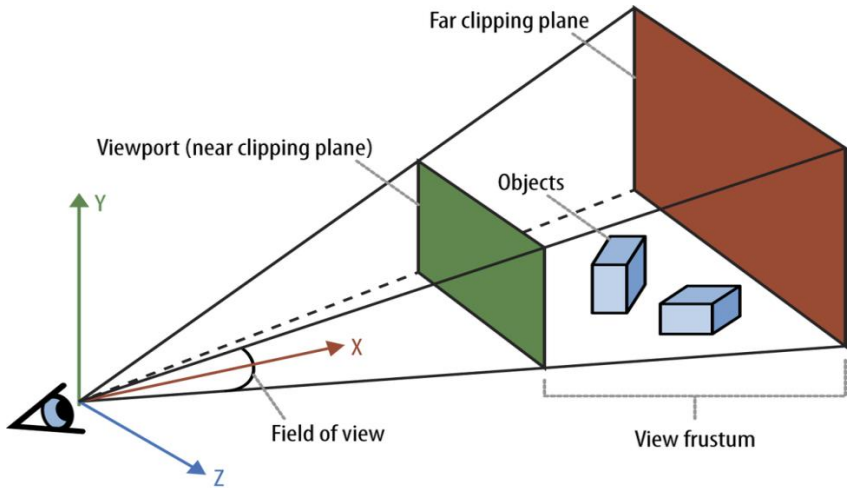

**Figure 12 Camera, viewport, and projection of real world objects onto 2D image plane** (*41*).

The projection calculations are held automatically in Augmented Reality (AR) platforms. After a crack or spall region is detected and accurately segmented from the scene, an image target is automatically created in the platform environment. The image targets work with feature-based 3D pose estimation using the calculated projection matrix (*42*). The projection matrix can be calculated by following the stereo camera calibration procedure provided by the headset manufacturers. In the calibration, camera intrinsic and extrinsic parameters such as camera focal length, location and orientation of the camera are estimated using the headset sensors gyroscope and head-position-tracker. After a successful calibration, simple proportioning of image pixel size to a known real-world dimension (camera offset from eye focus is known) is used to calculate the area of a spall or length of a crack. Figure show calibrated image targets in the AR platform.

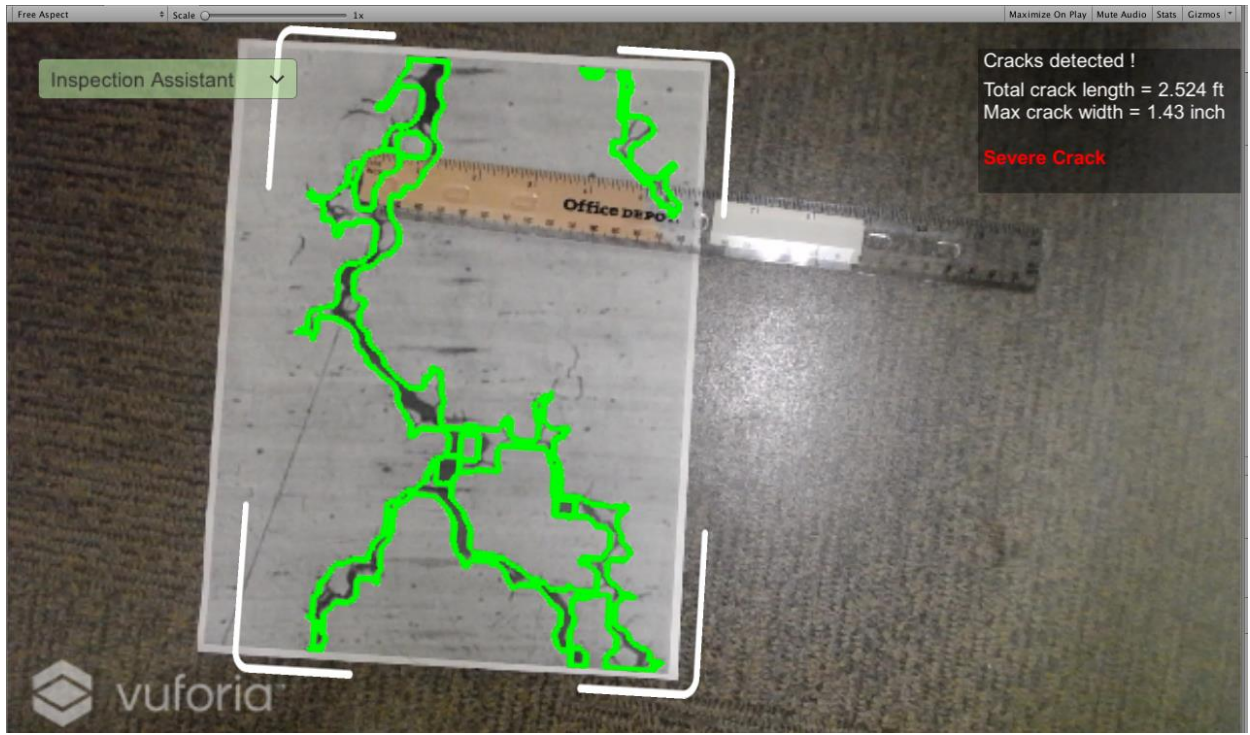

**Figure 13 Calibrating image targets using AR platform. (The calibration was performed in Unity using Vuforia library. A ruler was used to compare estimated maximum thickness.)**

### 3.6. Condition Assessment of Concrete Defects

the inspector would have the chance to investigate a certain defect in more detail if the condition information of the defect is shown to inspector in real-time. For example, when a crack condition is shown in the headset interface as "Severe Crack" according to AASHTO guideline, the inspector would like to perform a comprehensive crack assessment. This type of assistance to the inspector would lead to more objective and accurate inspection practice. The condition assessment methodology in this study aims to implement a quantified assessment procedure in which the limit values are interpreted from major inspection guidelines. The condition state limits and the recommended actions stated in FDOT, AASHTO, FHWA inspection guidelines are investigated. In AASHTO Bridge Inspection Manual (*27*), all elements have four defined condition states. The severity of multiple distress paths or deficiencies is defined in manual for each condition state with the general intent of the condition states as below. The feasible actions associated with each condition are also shown.

CS 1: Good  → do nothing/protect.
CS 2: Fair  → do nothing/protect/repair.

CS 3: Poor → do nothing/protect/repair/rehab
CS 4: Severe → do nothing/repair/rehab/replace

AASHTO manual provides somewhat quantifiable condition limits for cracking and delamination. Yet, other deterioration modes are mainly based on subjective decisions of visual inspection. The limits condition criteria for AASHTO are tabulated in Table 3.

**Table 3 Inspection condition limit criteria of AASHTO bridge deck elements**

| Defect | Hairline - Minor | Narrow - Moderate | Medium - Severe |
|---|---|---|---|
| Cracking | < 0.0625 inches (1/16") (1.6 mm) | 0.0625 – 0.125 inches (1/16"-1/8") (1.6 – 3.2 mm) | > 0.125 inches (1/8") (3.2 mm) |
| Spalls / Delamination | N/A | Spall less than 1 inch (25 mm) deep or less than 6 inches in diameter. No exposed rebar. | Spall greater than 1 inch (25 mm) deep or greater than 6 inches in diameter or exposed rebar |
| Cracking Density | Spacing Greater than 3.0 feet (0.33 m) | Spacing of 1.0 - 3.0 feet (0.33 – 1.0 m) | Spacing of less than 1 foot (0.33 m) |
| Efflorescence | N/A | Surface white without build-up or leaching | Heavy build-up with rust staining |

The condition assessment guides are used as reference in the MR system. An example implementation of the condition assessment methodology is shown in Figure 13.

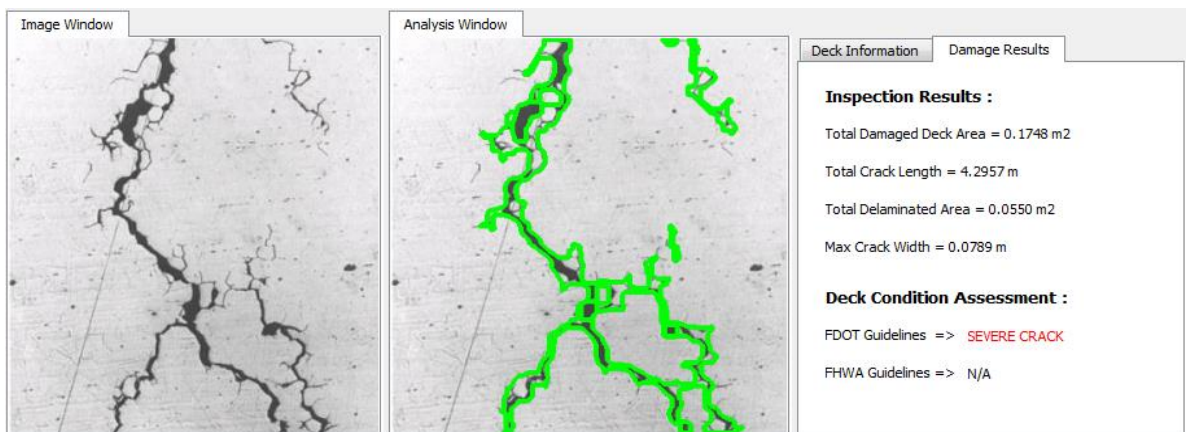

**Figure 14 Example implementation of the condition assessment methodology.**

## 4. CONCLUSIONS

This study aimed to integrate and demonstrate novel AI detection and segmentation algorithms into a MR framework by which a bridge inspector, for example, can benefit from this system during his/her routine inspection/assessment tasks. The inspector can analyze a damage in-real time and calculate its condition state without needing to perform any manual measurement. The methods described in the paper explain how a framework for collective human-AI intelligence can be created and how it can outperform the conventional or fully automated concrete inspections. The human-centered AI asks only minimal input from the human inspector and gets its predictions verified before finalizing a damage assessment task. This kind of a collaboration layer between human expert and AI is unique approach of this study. Furthermore, the AI system follows a semi-supervised learning approach and consistently improves itself with use of verified detection and segmentation data in re-training. The use of semi-supervised learning addresses successfully the problems of small data in AI training particularly encountered in damage detection applications where a comprehensive, publicly available image dataset is unavailable. This work aimed to achieve following scientific contributions with real life implementations for bridges and other structures:

- Current scientific approaches have employed various learning based methods for automatic detection of concrete defects while replacing human involvement in the process. However, the developed method aimed to merge engineer/inspector's expertise with AI assistance using a human-centered computing approach, thus yielding more reliable civil infrastructure visual assessment practice.

- In machine learning based approaches, the availability of training data is the most critical aspect of developing a reliable system with good accuracy in recognition. Yet, in infrastructure assessment, creating a large image dataset is particularly a challenging task. The proposed method therefore used an advanced data augmentation technique to generate synthetically sufficient amount of crack and spall images from the available image data.

- Utilizing Non-destructive Evaluation (NDE) methods effectively in bridge decision making has recently gained importance in the bridge management research with the growing number of vision based technologies for infrastructure inspections (i.e. camera based systems, unmanned aerial systems, infrared thermography, ground penetrating radar). This study proposed a method to collect more objective data for infrastructure management while also benefitting from inspectors' professional judgement. In the short-term, the proposed method can serve as an effective data collection method and in the long term, as the AI systems become more reliable approaches for infrastructure inspections, the proposed system will be a more feasible approach.

The AI assisted MR inspection framework presented will be expanded in many ways in a future study. First, a multichannel analysis method will be investigated in order fuse multiple sources of data (i.e. imagery data and infrared thermography). This new method will bring more capabilities such as detecting and analyzing subconcrete delamination and steel corrosion.

## 5. ACKNOWLEDGMENTS

AUTHOR CONFIRMATION

**The authors confirm contribution to the paper as follows: study conception and design: Enes Karaaslan, F. Necati Catbas, Ulas Bagci; data collection: Enes Karaaslan; analysis and interpretation of results: Enes Karaaslan, Ulas Bagci; draft manuscript preparation: Enes Karaaslan, F. Necati Catbas, Ulas Bagci. All authors reviewed the results and approved the final version of the manuscript.**

# REFERENCES

1. U.S. Department of Transportation Federal Highway Administation. *Deficient Bridges by Highway System*. 2017.

2. Catbas, F., S. Ciloglu, A. Aktan, F. N. Catbas, S. K. Ciloglu{, and A. E. Aktan{. Strategies for Load Rating of Infrastructure Populations: A Case Study on T-Beam Bridges. 2017. https://doi.org/10.1080/15732470500031008.

3. Brown, M. C., J. P. Gomez, M. L. Hammer, and J. M. Hooks. *Long-Term Bridge Performance High Priority Bridge Performance Issues*. McLean, VA, 2014.

4. Behzadan, A. H., S. Dong, and V. R. Kamat. Augmented Reality Visualization: A Review of Civil Infrastructure System Applications. *Advanced Engineering Informatics*, Vol. 29, No. 2, 2015, pp. 252–267. https://doi.org/10.1016/j.aei.2015.03.005.

5. Mihelj, M., D. Novak, and S. Beguš. *Virtual Reality Technology and Applications*. 2014.

6. Azuma, R., R. Behringer, S. Feiner, S. Julier, and B. Macintyre. Recent Advances in Augmented Reality. *IEEE Computer Graphics and Applications*, Vol. 2011, No. December, 2001, pp. 1–27. https://doi.org/10.4061/2011/908468.

7. Sutherland, I. E. The Ultimate Display. 1965.

8. Milgram, P., and F. Kishino. Taxonomy of Mixed Reality Visual Displays. *IEICE Transactions on Information and Systems*, Vol. E77–D, No. 12, 1994, pp. 1321–1329. https://doi.org/10.1.1.102.4646.

9. Windows Mixed Reality. What Is Mixed Reality ?

10. Coutrix, C., and L. Nigay. Mixed Reality: A Model of Mixed Interaction. *Proceedings of the working conference on Advanced visual interfaces - AVI '06*, 2006, pp. 43–50. https://doi.org/10.1145/1133265.1133274.

11. LaViola, J., E. Kruijff, D. Bowman, and I. Poupyrev. *3D User Interfaces: Theory and Practice, Second Edition*. 2017.

12. Kamat, V. R., and S. El-Tawil. Evaluation of Augmented Reality for Rapid Assessment of Earthquake-Induced Building Damage. *Journal of Computing in Civil Engineering*, Vol. 21, No. 5, 2007, pp. 303–310. https://doi.org/10.1061/(ASCE)0887-3801(2007)21:5(303).

13. Behzadan, A. H., and V. R. Kamat. Georeferenced Registration of Construction Graphics in Mobile Outdoor Augmented Reality. *Journal of Computing in Civil Engineering*, Vol. 21, No. 4, 2007, pp. 247–258. https://doi.org/10.1061/(ASCE)0887-3801(2007)21:4(247).

14. Bae, H., M. Golparvar-Fard, and J. White. High-Precision Vision-Based Mobile Augmented Reality System for Context-Aware Architectural, Engineering, Construction and Facility Management (AEC/FM) Applications. *Visualization in Engineering*, Vol. 1,

No. 1, 2013, p. 3. https://doi.org/10.1186/2213-7459-1-3.

15. Ioannis, B. Mixed Reality Constructs a New Frontier for Maintaining the Built Environment. *Proceedings of the Institution of Civil Engineers - Civil Engineering*, Vol. 170, No. 2, 2017, p. 53. https://doi.org/10.1680/jcien.2017.170.2.53.

16. Moreu, F., B. Bleck, S. Vemuganti, D. Rogers, and D. Mascarenas. Augmented Reality Tools for Enhanced Structural Inspection. No. 2, 2017, pp. 3124–3130.

17. Microsoft. The Leader in Mixed Reality Technology | HoloLens. *Microsoft*.

18. Koch, C., K. Georgieva, V. Kasireddy, B. Akinci, and P. Fieguth. A Review on Computer Vision Based Defect Detection and Condition Assessment of Concrete and Asphalt Civil Infrastructure. *Advanced Engineering Informatics*, Vol. 29, No. 2, 2015, pp. 196–210. https://doi.org/10.1016/j.aei.2015.01.008.

19. Yokoyama, S., and T. Matsumoto. Development of an Automatic Detector of Cracks in Concrete Using Machine Learning. *Procedia Engineering*, Vol. 171, 2017, pp. 1250–1255. https://doi.org/10.1016/j.proeng.2017.01.418.

20. Jahanshahi, M. R., and S. F. Masri. Adaptive Vision-Based Crack Detection Using 3D Scene Reconstruction for Condition Assessment of Structures. *Automation in Construction*, Vol. 22, 2012, pp. 567–576. https://doi.org/10.1016/j.autcon.2011.11.018.

21. Adhikari, R. S., O. Moselhi, and A. Bagchi. Image-Based Retrieval of Concrete Crack Properties for Bridge Inspection. *Automation in Construction*, Vol. 39, 2014, pp. 180–194. https://doi.org/10.1016/j.autcon.2013.06.011.

22. German, S., I. Brilakis, and R. Desroches. Rapid Entropy-Based Detection and Properties Measurement of Concrete Spalling with Machine Vision for Post-Earthquake Safety Assessments. *Advanced Engineering Informatics*, Vol. 26, No. 4, 2012, pp. 846–858. https://doi.org/10.1016/j.aei.2012.06.005.

23. Zaurin, R., T. Khuc, F. N. Catbas, and F. Asce. Hybrid Sensor-Camera Monitoring for Damage Detection: Case Study of a Real Bridge. *Journal of Bridge Engineering*, Vol. 21, No. 6, 2015, pp. 1–27. https://doi.org/10.1061/(ASCE)BE.1943.

24. Hiasa, S., F. N. Catbas, M. Matsumoto, and K. Mitani. Monitoring Concrete Bridge Decks Using Infrared Thermography with High Speed Vehicles. *Structural Monitoring and Maintenance, An International Journal*, Vol. 3, No. 3, 2016, pp. 277–296. https://doi.org/http://dx.doi.org/10.12989/smm.2016.3.3.277.

25. Hiasa, S. *Investigation of Infrared Thermography for Subsurface Damage Detection of Concrete Structures*. Electronic Theses and Dissertations. Paper 5063. <http://stars.library.ucf.edu/etd/5063>, 2016.

26. Xie, D., L. Zhang, and L. Bai. Deep Learning in Visual Computing and Signal Processing. 2017, pp. 1–18. https://doi.org/10.1155/2017/1320780.

27. AASHTO. Guide Manual for Bridge Element Inspection. *Bridge Element Inspection Manual*, 2011, p. 172.

28. FHWA. National Bridge Inspection Standards Regulations (NBIS). *Federal Register*, Vol. 69, No. 239, 2004, pp. 15–35.

29. FDOT. *Florida DOT Bridge Inspection Field Guide*. 2016.

30. Prasanna, P., K. J. Dana, N. Gucunski, B. B. Basily, H. M. La, R. S. Lim, and H. Parvardeh. Automated Crack Detection on Concrete Bridges. *IEEE Transactions on Automation Science and Engineering*, Vol. 13, No. 2, 2016, pp. 591–599. https://doi.org/10.1109/TASE.2014.2354314.

31. Maeda, H., Y. Sekimoto, T. Seto, T. Kashiyama, and H. Omata. Road Damage Detection Using Deep Neural Networks with Images Captured Through a Smartphone. 2015, pp. 4–6.

32. Yang, L., B. Li, W. Li, Z. Liu, G. Yang, and J. Xiao. A Robotic System towards Concrete Structure Spalling and Crack Database. *2017 IEEE International Conference on Robotics and Biomimetics (ROBIO), Robotics and Biomimetics (ROBIO), 2017 IEEE International Conference on*. 1276. https://login.ezproxy.net.ucf.edu/login?auth=shibb&url=http://search.ebscohost.com/login.aspx?direct=true&db=edseee&AN=edseee.8324593&site=eds-live&scope=site.

33. UCF Advanced Research Computing Center. Newton Visualization Cluster. https://arcc.ist.ucf.edu.

34. Liu, W., D. Anguelov, D. Erhan, C. Szegedy, S. Reed, C. Y. Fu, and A. C. Berg. SSD: Single Shot Multibox Detector. *Lecture Notes in Computer Science (including subseries Lecture Notes in Artificial Intelligence and Lecture Notes in Bioinformatics)*, Vol. 9905 LNCS, 2016, pp. 21–37. https://doi.org/10.1007/978-3-319-46448-0_2.

35. Sandler, M., A. Howard, M. Zhu, A. Zhmoginov, and L.-C. Chen. MobileNetV2: Inverted Residuals and Linear Bottlenecks. 2018. https://doi.org/10.1134/S0001434607010294.

36. Simonyan, K., and A. Zisserman. Very Deep Convolutional Networks for Large-Scale Image Recognition. *CoRR*, Vol. abs/1409.1, 2014.

37. LaLonde, R., and U. Bagci. Capsules for Object Segmentation. No. Midl, 2018, pp. 1–9.

38. Badrinarayanan, V., A. Kendall, and R. Cipolla. SegNet: A Deep Convolutional Encoder-Decoder Architecture for Image Segmentation. *IEEE Transactions on Pattern Analysis and Machine Intelligence*, Vol. 39, No. 12, 2017, pp. 2481–2495. https://doi.org/10.1109/TPAMI.2016.2644615.

39. Hill, M. Overview of Human-Computer Collaboration. Vol. 8, No. June, 1995, pp. 67–81.

40. Chen, S., L. Liang, W. Liang, and H. Foroosh. 3D Pose Tracking with Multitemplate

Warping and SIFT Correspondences. *IEEE Transactions on Circuits and Systems for Video Technology*, Vol. 26, No. 11, 2016, pp. 2043–2055. https://doi.org/10.1109/TCSVT.2015.2452782.

41. Parisi, T. *Programming 3D Applications with HTML5 and WebGL*. 2014.

42. Marchand, E., H. Uchiyama, F. Spindler, E. Marchand, H. Uchiyama, and F. Spindler. Pose Estimation for Augmented Reality : A Hands-on Survey Pose Estimation for Augmented Reality : A Hands-on Survey. Vol. 22, No. 12, 2016, pp. 2633–2651. https://doi.org/10.1109/TVCG.2015.2513408.